\documentclass[letterpaper, 10 pt, conference]{ieeeconf}

\IEEEoverridecommandlockouts
\usepackage{amsmath,amsfonts}
\usepackage{algorithmic}
\usepackage{algorithm}
\usepackage{multirow}
\usepackage{subcaption}
\usepackage{xcolor}
\usepackage{makecell}
\usepackage{array}

\usepackage{textcomp}
\usepackage{stfloats}
\usepackage{url}
\usepackage{verbatim}
\usepackage{graphicx}
\usepackage{tabularx}
\usepackage{cite}
\usepackage{optidef}
\usepackage{booktabs}    % For professional-quality tables (\toprule, \midrule, etc.)
\usepackage[colorlinks, linkcolor=red, anchorcolor=green, citecolor=blue ]{hyperref}         
\title{\vspace{0pt}%
\centering\LARGE \bf ADM-DP: Adaptive Dynamic Modality Diffusion Policy through Vision-Tactile-Graph Fusion for Multi-Agent Manipulation}

\author{Enyi Wang, Wen Fan and Dandan Zhang%
\thanks{Enyi Wang, Wen Fan, and Dandan Zhang are with the Department of Bioengineering, Imperial-X Initiative, Imperial College London, London, United Kingdom. Corresponding: d.zhang17@imperial.ac.uk}%
}

\begin{document}
\maketitle

\begin{abstract}
Multi-agent robotic manipulation remains challenging due to the combined demands of coordination, grasp stability, and collision avoidance in shared workspaces. To address these challenges, we propose the Adaptive Dynamic Modality Diffusion Policy (ADM-DP), a framework that integrates vision, tactile, and graph-based (multi-agent pose) modalities for coordinated control.  ADM-DP introduces four key innovations. First, an enhanced visual encoder merges RGB and point-cloud features via Feature-wise Linear Modulation (FiLM) modulation to enrich perception. Second, a tactile-guided grasping strategy uses Force-Sensitive Resistor (FSR) feedback to detect insufficient contact and trigger corrective grasp refinement, improving grasp stability.  Third, a graph-based collision encoder leverages shared \emph{tool center point} (TCP) positions of multiple agents as structured kinematic context to maintain spatial awareness and reduce inter-agent interference. Fourth, an Adaptive Modality Attention Mechanism (AMAM) dynamically re-weights modalities according to task context, enabling flexible fusion.
For scalability and modularity, a decoupled training paradigm is employed in which agents learn independent policies while sharing spatial information. This maintains low interdependence between agents while retaining collective awareness. Across seven multi-agent tasks, ADM-DP achieves 12-25\% performance gains over state-of-the-art baselines. Ablation studies show the greatest improvements in tasks requiring multiple sensory modalities, validating our adaptive fusion strategy and demonstrating its robustness for diverse manipulation scenarios.   \url{https://Enyi-Bean.github.io/ADM-DP/}
\end{abstract}

% \begin{IEEEkeywords}
% Multimodal Fusion, Tactile Sensing, Diffusion Policy, Multi-agent Manipulation
% \end{IEEEkeywords}

% \textbf{Keywords—}Multimodal fusion, tactile sensing, diffusion policy, multi-agent manipulation.

\section{Introduction}

Imitation learning has become a powerful paradigm for robotic manipulation, enabling robots to acquire complex skills from demonstrations without explicit programming~\cite{hussein2017imitation}. Recent advances in Diffusion-based policies have revolutionized this field: Diffusion Policy (DP)~\cite{chi2023diffusion} models actions as conditional denoising processes, and 3D Diffusion Policy~\cite{ze20243d} leverages point clouds for improved spatial reasoning. Subsequent work has explored richer visual representations~\cite{ma2025cdp,wu2025afforddp}, while flow-matching alternatives such as Flow Policy~\cite{zhang2025flowpolicy,lipman2022flow} achieve faster inference via straight-through trajectory generation. However, these vision-centric methods are predominantly designed for single-agent settings; extending them to multi-agent manipulation exposes unresolved challenges in coordination, collision avoidance, and multi-sensory fusion.

% \textbf{The Multi-Agent Coordination Challenge.} 

Current multi-agent systems face a critical trade-off between scalability and coordination effectiveness \cite{chen2025multi}. Centralized approaches that jointly process all agents' observations suffer from exponential state-space growth. Recent work has explored decoupled training paradigms, with Jiang et al.~\cite{jiang2025rethinking} proposing a decoupled interaction framework for bimanual tasks and RoboFactory~\cite{qin2025robofactory} extending this to multi-agent scenarios. While these methods achieve better scalability, they lack explicit collision awareness mechanisms. Graph neural networks (GNNs) show promise for multi-robot coordination~\cite{lai2025roboballet}, yet require explicit scene modeling that limits task generalization. The recent KStar Diffuser~\cite{lv2025spatial} constructs comprehensive joint-level spatio-temporal graphs, but includes many task-irrelevant nodes (e.g., base joints) that increase computational overhead without improving end-effector coordination where collisions actually occur. This suggests a key insight: multi-agent systems require lightweight coordination which focuses on task-critical interactions.

\begin{figure}
\centering
\captionsetup{font=footnotesize,labelsep=period}
\includegraphics[width=\columnwidth]{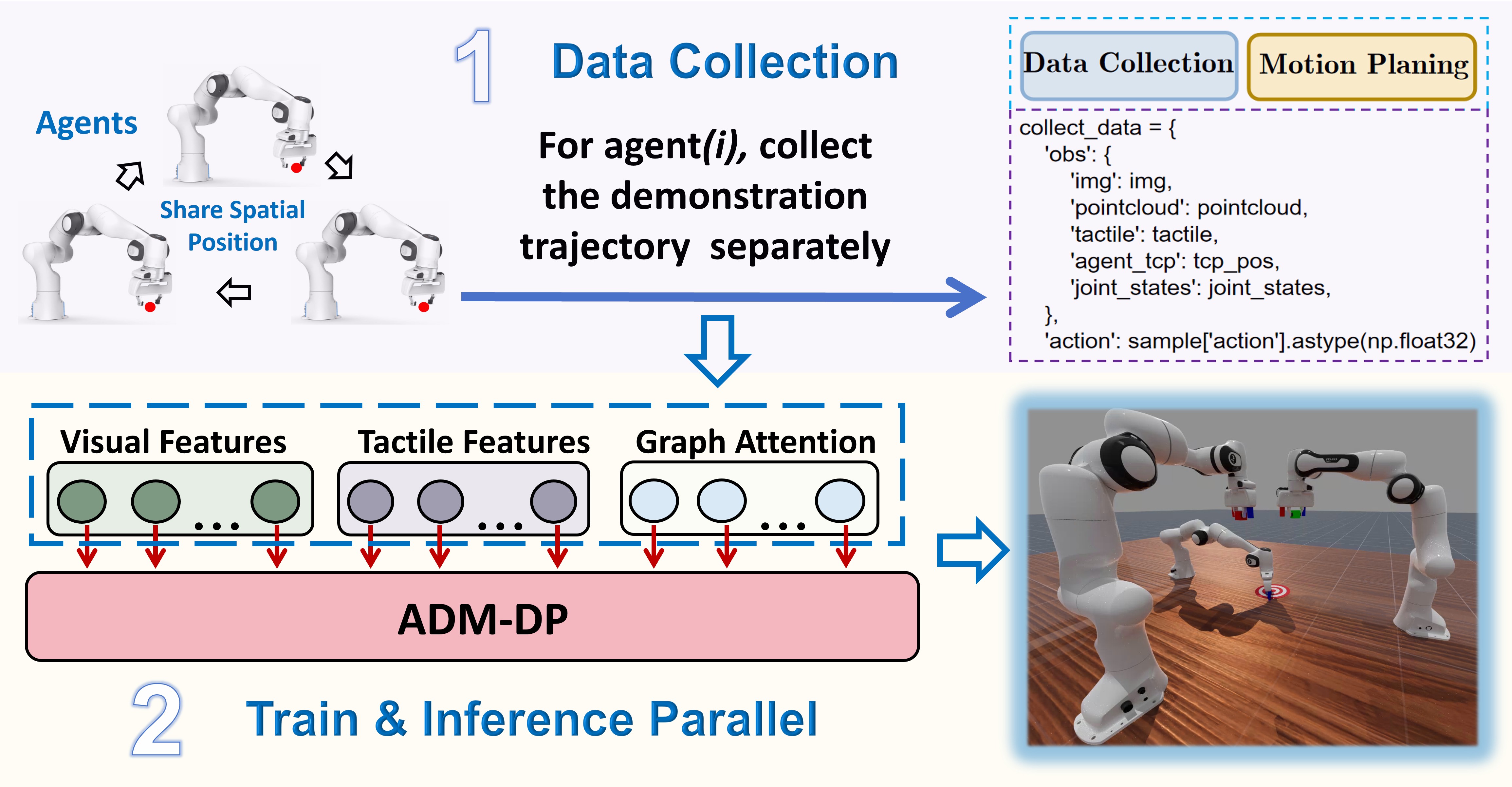}
\caption{The overview pipline of our method framework.}
\label{fig:pipeline}
 \vspace{-0.6cm}
\end{figure}

% \textbf{The Static Fusion Problem in Multi-Modal Learning.} 

Furthermore, the demand upon coordination highlights only one side of the challenge. Equally important is how multi-agent systems handle multi-modal sensory input, which can undermine both efficiency and robustness. Inspired by bionic behaviour, humans naturally modulate sensory attention during manipulation, relying on vision for approach, touch for contact verification, and spatial awareness for coordination. Yet most of current multi-modal methods~\cite{huang20243d,xue2025reactive,gu2025tactilealoha} employ static fusion strategies, only concatenating or uniformly weighting modalities regardless of task phase. This creates fundamental inefficiencies across all sensory channels: tactile signals are zero during approach phases, providing no information yet still being encoded as high-dimensional features that inject noise into the network; spatial awareness for collision avoidance is unnecessary when agents are distant but becomes critical when proximate; even visual features may be redundant during stable grasping when tactile feedback should dominate. 

Advanced tactile-integrated architectures like Reactive Diffusion Policy's dual-frequency design~\cite{xue2025reactive} or 3D-ViTac's unified tactile point cloud representation~\cite{huang20243d} continue processing these zero-valued or irrelevant channels throughout execution. Methods like Bi-Touch~\cite{lin2023bi} assume all modalities are `always useful', degrading sample efficiency when certain readings are meaningless noise. The core limitation is clear: without dynamic modality weighting that adapts to task context, policies waste computational resources on redundant sensing information irrelevant to current task state while potentially allowing their noise to corrupt useful signals from task-critical modalities.

% \textbf{The Grasp Stability Challenge Under Occlusion.} 

This limitation becomes especially critical in multi-agent manipulation, where mutual occlusions and workspace interference amplify the impact of poorly integrated sensory inputs, leading to unstable grasps. While tactile sensing provides rich contact information, existing methods~\cite{xue2025reactive, lin2023bi, huang20243d} treat it as supplementary observation rather than leveraging it for active control. The challenge extends beyond sensor integration, it requires data collection strategies that teach policies to recognize insufficient grasps through tactile signatures and execute corrective actions, transforming tactile feedback from passive sensing to active refinement signal.

These interconnected challenges motivate our core insight: multi-agent manipulation requires adaptive sensory fusion that dynamically adjusts to task phases, not static architectures that process all modalities uniformly. To achieve these objectives, we define the \textbf{modality} used in this work as any structured information source that contributes complementary perspectives for cooperative manipulation across multiple agents, including sensory signals (e.g., vision, tactile) and relational encodings (e.g., graph-based spatial awareness). Accordingly, we propose \textbf{ADM-DP} (Adaptive Dynamic Modality Diffusion Policy), a framework illustrated in Fig.~\ref{fig:pipeline}. 
In summary, we introduce three key contributions:

%\textbf{(1) Tactile-guided grasping strategy.} We design a data-collection protocol that deliberately injects shallow-to-deep grasp adjustments into 
% 30\% of demonstrations. In these cases, initial shallow grasps with partial Force-Sensitive Resistor (FSR) activation are followed by corrective deepening motions. This exposure enables the policy to link tactile signatures with insufficient contact and to execute refinement actions, thereby transforming tactile sensing from passive observation into active control. In combination with the decoupled training paradigm, in which agents learn independent policies while sharing \emph{tool center point} TCP positions, the strategy supports scalable learning while maintaining effective inter-agent coordination.

\textbf{(1) Tactile-guided grasping strategy.} We design a data-collection protocol that deliberately includes shallow-to-deep grasp refinements in 30\% of demonstrations. In these trajectories, an initial shallow grasp, indicated by weak Force-Sensitive Resistor (FSR)-based tactile signals and incomplete contact, is followed by a corrective deepening motion. This exposure enables the policy to associate tactile signatures with insufficient contact and to execute refinement actions, thereby elevating tactile sensing from passive observation to active control. Combined with the decoupled training paradigm, in which agents learn independent policies while conditioning on shared TCP position information, this strategy supports scalable learning while preserving effective inter-agent coordination.

\textbf{(2) Multi-modal Encoding for Complementary Sensing:} ADM-DP processes each modality through specialized encoders designed for their unique characteristics. We enhance visual perception by combining RGB semantics with point cloud geometry via FiLM modulation, providing robust 3D understanding despite occlusions. Tactile signals from FSR arrays are encoded with spatial positions to preserve contact patterns crucial for grasp stability. Graph Attention Networks (GAT)~\cite{velivckovic2017graph} process shared TCP positions in graph structural format to enable lightweight collision avoidance between multiple agents without modeling entire kinematic chains. Each encoder extracts task-critical features while minimizing computational overhead.

\textbf{(3) Dynamic Modality Fusion via AMAM:} Unlike static fusion methods that process all modalities uniformly, our proposed AMAM dynamically allocates attention based on task context through learnable importance weights. With entropy regularization preventing both uniform averaging and modality collapse, AMAM learns to suppress tactile noise during approach, amplify it during contact, and activate spatial awareness when agents converge. This adaptive fusion enables policies to automatically adjust sensory priorities throughout task execution without manual phase detection or switching.
By addressing the fundamental limitation of static fusion and introducing adaptive mechanisms for multi-agent coordination, ADM-DP advances toward robotic systems that dynamically modulate their sensory focus based on task demands, knowing when to prioritize vision, when to rely on touch, and when to maintain spatial awareness.

\section{Methodology}

We address the problem of multi-agent robotic manipulation where $n$ agents must coordinate to complete complex tasks. Let $\mathcal{A} = \{\mathcal{A}_1, ..., \mathcal{A}_n\}$ represent the entire action space where $\mathcal{A}_i$ is the action space for specific agent $i$. The observation space for each agent $i$ consists of both local and shared components: $\mathcal{O}_i = \{\mathcal{O}_i^{\text{local}}, \mathcal{O}^{\text{shared}}\}$. The local observations $\mathcal{O}_i^{\text{local}} = \{I_i, P_i, T_i, q_i, L_i\}$ include RGB images $I_i \in \mathbb{R}^{H \times W \times 3}$, point clouds $P_i \in \mathbb{R}^{N \times 6}$, tactile readings $T_i \in \mathbb{R}^{32}$ from FSR sensors, joint states $q_i \in \mathbb{R}^{d_q}$, and a language instruction $L_i$ specifying the agent's sub-task. The shared observations $\mathcal{O}^{\text{shared}} = \{p_1^{\text{tcp}}, ..., p_n^{\text{tcp}}\}$ contain the end-effector TCP positions $p_i^{\text{tcp}} \in \mathbb{R}^3$ of all agents, enabling collision-aware coordination. Our goal is to learn a set of decoupled policies $\{\pi_1, ..., \pi_n\}$ where each policy $\pi_i: \mathcal{O}_i \rightarrow \mathcal{A}_i$ maps agent $i$'s observations to actions. We present our approach in the following sections: the decoupled training paradigm (Sec. II-A), the ADM-DP architecture (Sec. II-B), and tactile-guided grasping strategy (Sec. II-C).

\subsection{Decoupled Training and Evaluation Paradigm}

Traditional approaches of bimanual or multi-agent robotic manipulation often employ a centralized policy $\pi: \mathcal{O}_1 \times ... \times \mathcal{O}_n \rightarrow \mathcal{A}_1 \times ... \times \mathcal{A}_n$ that jointly processes all agents' observations and outputs all actions simultaneously as well. However, this approach suffers from several limitations: exponential growth in observation and action spaces with increasing agents, difficulty in generalizing to different numbers of agents, and high sample complexity for training.

In contrast, recent works have explored decoupled approaches for multi-arm manipulation. Jiang et al.~\cite{jiang2025rethinking} proposed a decoupled interaction framework for bimanual tasks, while RoboFactory~\cite{qin2025robofactory} demonstrated that training separate policies for each agent can effectively scale to multi-agent scenarios. Building upon these decoupled training strategies, we adopt a similar paradigm with an important extension: our agents share TCP positions with each other in addition to their local observations to enhance coordination. Therefore, we adopt a decoupled training paradigm. During training, we learn independent policies for each agent:
\begin{equation}
\text{Training: } \pi_i: \mathcal{O}_i \rightarrow \mathcal{A}_i, \quad i \in \{1, ..., n\}
\end{equation}
where each policy $\pi_i$ is optimized separately using demonstrations collected via motion planning with randomized initializations. Each agent $i$ observes $\mathcal{O}_i = \{\mathcal{O}_i^{\text{local}}, \mathcal{O}^{\text{shared}}\}$, where the shared component contains all agents' TCP positions for spatial awareness. At evaluation time, the trained policies execute in parallel:
\begin{equation}
\text{Evaluation: } a_i^t = \pi_i(\mathcal{O}_i^t), \quad \forall i \in \{1, ..., n\}
\end{equation}
where each agent $\mathcal{A}_i$ independently generates actions $a_i^t$ based on its current observations $\mathcal{O}_i^t$, with TCP positions updated in real-time from all agents.

This decoupled approach significantly reduces training complexity as each policy only needs to learn single-agent behaviors rather than the exponentially larger joint action space. Moreover, it enables modular deployment where agents can be added or removed without retraining the entire system.

\subsection{ADM-DP Architecture}

As illustrated in Fig.~\ref{fig:policy}, ADM-DP processes multi-modal observations through specialized encoders and dynamically fuses them via AMAM, which consists of three main components: (1) multi-modal encoders for vision, tactile, and graph modalities; (2) AMAM fusion module; and (3) diffusion-based action decoder:

\begin{figure*}
\centering
\captionsetup{font=footnotesize,labelsep=period}
\includegraphics[width=\textwidth]{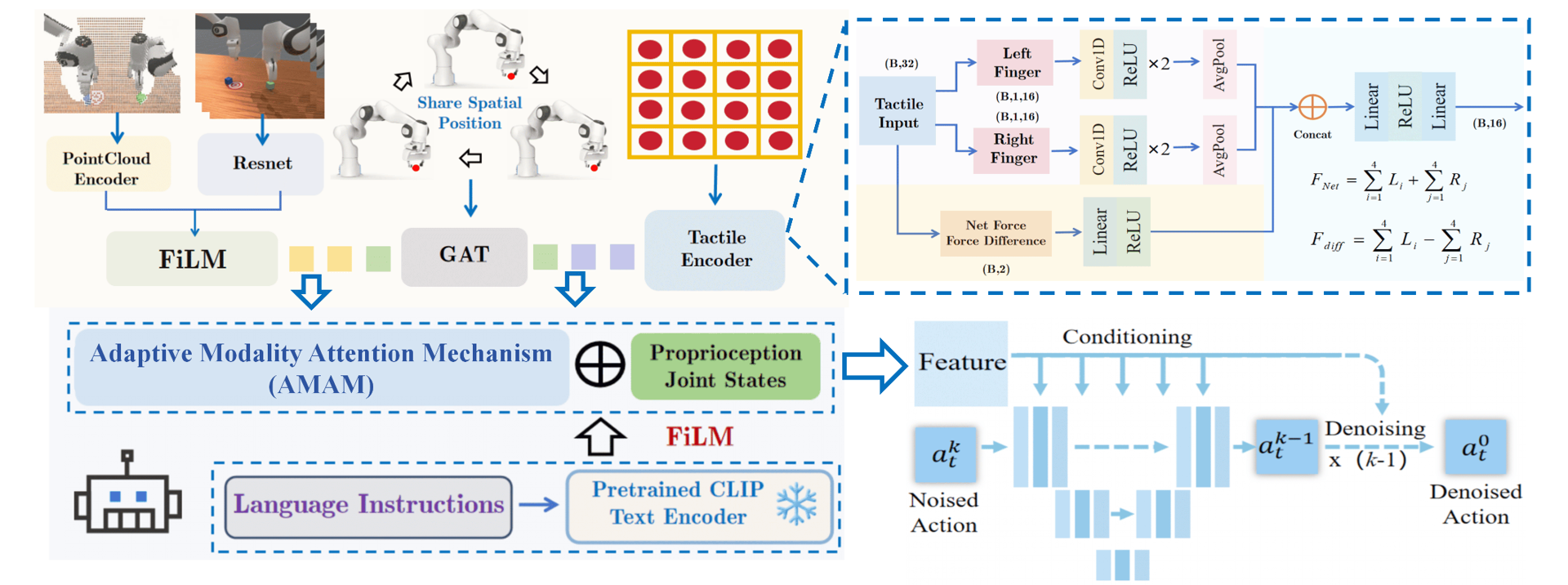}
\caption{Overview of ADM-DP architecture. Multi-modal observations (vision, tactile, graph) are processed through specialized encoders and dynamically fused via AMAM. The fused features are conditioned on language instructions through FiLM~\cite{perez2018film} modulation before guiding the diffusion process for action generation.}
\label{fig:policy}
 \vspace{-0.1cm}
\end{figure*}

\subsubsection{Multi-modal Encoders}

\textbf{(1) Vision Encoding:} 
We process visual inputs through complementary RGB and point cloud pathways. RGB images $I_i \in \mathbb{R}^{H \times W \times 3}$ are encoded via ResNet~\cite{he2016deep} to extract semantic features $f_{img} = \text{ResNet}(I_i) \in \mathbb{R}^{512}$. Point clouds undergo preprocessing including workspace cropping and downsampling to $N=1024$ points while preserving color information, resulting in $P_i \in \mathbb{R}^{1024 \times 6}$. These are processed through a modified PointNet~\cite{qi2017pointnet} architecture where we remove T-Net and BatchNorm layers following insights from DP3~\cite{ze20243d}, yielding geometric features $f_{pc} = \text{PointNet}(P_i) \in \mathbb{R}^{1024}$. To leverage the complementary strengths of both modalities, semantic understanding from RGB and precise 3D geometry from point clouds, we integrate them using FiLM~\cite{perez2018film} modulation:
\begin{equation}
f_v = \gamma(f_{pc}) \odot f_{img} + \beta(f_{pc})
\end{equation}
where $\gamma(\cdot)$ and $\beta(\cdot)$ are learned affine transformations. This fusion strategy allows geometric features to adaptively modulate semantic features, producing enhanced visual representations $f_v \in \mathbb{R}^{512}$ that are more robust to visual ambiguities and occlusions common in multi-agent scenarios.

\textbf{(2) Tactile Encoding:}
Each gripper is equipped with 4×4 FSR sensors on both fingertips, providing tactile readings $T_i \in \mathbb{R}^{32}$. To preserve spatial structure crucial for grasp adjustment, we incorporate positional encoding for each taxel. The tactile encoder first reshapes the input into $T_i \in \mathbb{R}^{2 \times 4 \times 4}$ (two fingers, each with 4×4 grid), applies log-normalization for stability, and concatenates 2D grid positions $(x, y) \in [-1, 1]^2$ to each taxel reading. Per-finger features are extracted through 1D convolutions with adaptive pooling. Additionally, we compute contact dynamics including (1) resultant force (sum of all taxel readings), and (2) differential force between fingers (indicating grasp balance) for each finger to track contact point locations. These physical features, combined with the convolutional features, are fused through a feedforward network to produce $f_t \in \mathbb{R}^{64}$, enabling the model to understand both fine-grained contact patterns and global force distributions critical for stable grasping.

\textbf{(3) Graph-based Collision Encoding:}
To enable collision-aware coordination, we encode the shared TCP positions $\mathcal{O}^{\text{shared}} = \{p_1^{\text{tcp}}, ..., p_n^{\text{tcp}}\}$ using a Graph Attention Network (GAT)~\cite{velivckovic2017graph}. Each TCP position forms a node in a fully connected graph with edge weights inversely proportional to distances:
\begin{equation}
e_{ij} = \frac{1}{\|p_i^{\text{tcp}} - p_j^{\text{tcp}}\|_2 + \epsilon}
\end{equation}
The GAT aggregates spatial relationships through attention mechanisms, producing graph features $f_g \in \mathbb{R}^{64}$ that capture proximity-aware inter-agent relationships.

\begin{figure}
\centering
\captionsetup{font=footnotesize,labelsep=period}
\includegraphics[width=1\columnwidth]{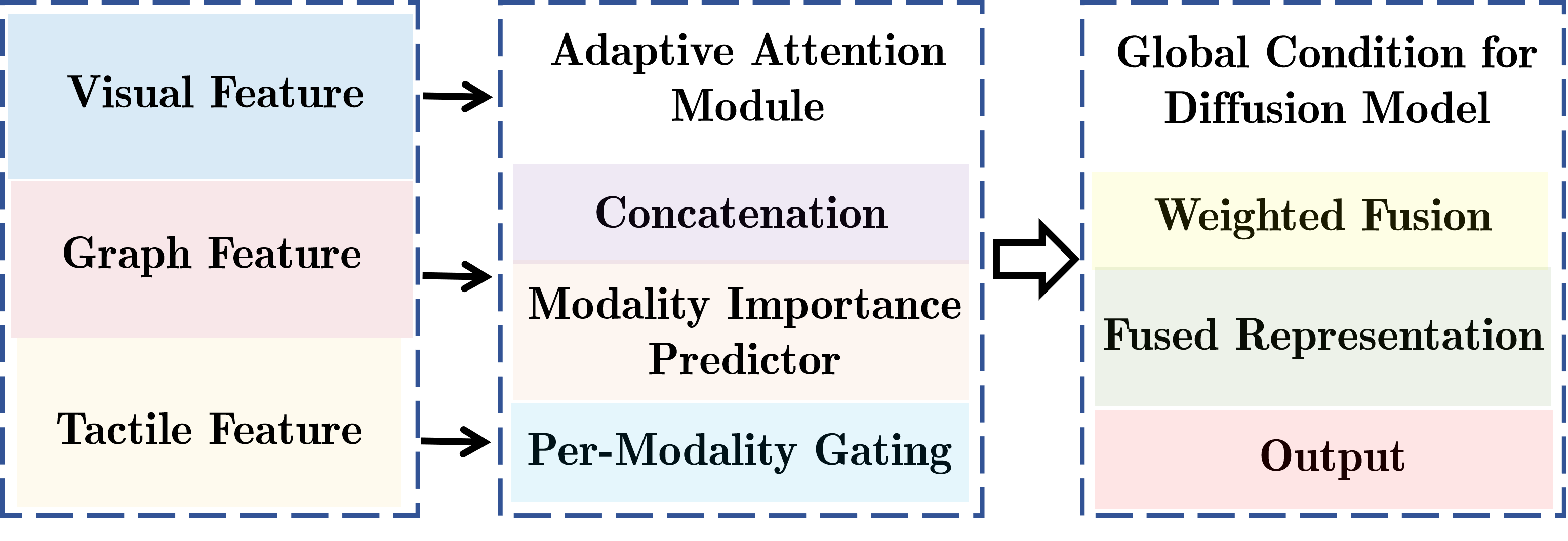}
\caption{Adaptive Modality Attention Mechanism (AMAM) dynamically allocates importance weights to vision, tactile, and graph modalities based on task context.}
\label{fig:cimm}
 \vspace{-0.3cm}
\end{figure}

\subsubsection{AMAM Fusion Module}

\textbf{(1) AMAM:}
Different task phases require adaptive modality priorities: vision dominates during approaching before contact, tactile becomes crucial during grasping, and graph awareness intensifies when agents are in close proximity. As shown in Fig.~\ref{fig:cimm}, our AMAM module dynamically allocates importance weights to each modality based on the current context. Given the encoded features $\{f_v, f_t, f_g\}$ in terms of vision, tactile and graph, AMAM computes importance weights through a gated attention mechanism:
\begin{equation}
\alpha = \text{softmax}(\text{MLP}([f_v; f_t; f_g]) / \tau)
\end{equation}
where $\tau$ is a temperature parameter and MLP is a multi-layer perceptron. The fused feature is computed as:
\begin{equation}
f_{vtg} = \alpha_v \cdot f_v + \alpha_t \cdot f_t + \alpha_g \cdot f_g
\end{equation}

To prevent weight collapse where all modalities receive equal attention, we introduce an entropy regularization term:
\begin{equation}
\mathcal{L}_{reg} = -\lambda \sum_{m \in \{v,t,g\}} \alpha_m \log(\alpha_m)
\end{equation}
This encourages the model to make decisive modality selections rather than uniform averaging.

\textbf{(2) Conditional Feature Integration:}
The fused multi-modal features are concatenated with proprioceptive joint states: $f_{obs} = [f_{vtg}; q_i]$. Language instructions $L_i$ are encoded using a frozen CLIP~\cite{radford2021learning} text encoder to obtain $f_l = \text{CLIP}(L_i)$. The final conditioning feature is obtained through FiLM modulation:
\begin{equation}
f_{cond} = \gamma(f_l) \odot f_{obs} + \beta(f_l)
\end{equation}
This conditioning vector $f_{cond}$ guides the diffusion process for action generation, enabling task-specific behaviors while maintaining multi-modal awareness.

\subsubsection{Diffusion-based Action Generation}
We employ a conditional diffusion model to generate action trajectories given the multi-modal conditioning features. Following the DDPM framework~\cite{ho2020denoising}, we define a forward diffusion process that gradually adds Gaussian noise to action trajectories over $T$ timesteps:
\begin{equation}
q(a^k|a^{k-1}) = \mathcal{N}(a^k; \sqrt{1-\beta_k}a^{k-1}, \beta_k\mathbf{I})
\end{equation}
where $a^0$ represents the clean action trajectory, $a^T$ is pure Gaussian noise, and $\{\beta_k\}_{k=1}^T$ is a variance schedule.

The reverse process learns to denoise actions conditioned on the observation features $f_{cond}$:
\begin{equation}
p_\theta(a^{k-1}|a^k, f_{cond}) = \mathcal{N}(a^{k-1}; \mu_\theta(a^k, k, f_{cond}), \sigma_k^2\mathbf{I})
\end{equation}
where $\mu_\theta$ is parameterized by a U-Net architecture that takes the noisy action, timestep, and conditioning features as input.

During training, we optimize the network to predict the noise added at each timestep:
\begin{equation}
\mathcal{L}_{diff} = \mathbb{E}_{k,\epsilon,a^0} \left[ \|\epsilon - \epsilon_\theta(a^k, k, f_{cond})\|^2 \right]
\end{equation}
where $\epsilon \sim \mathcal{N}(0, \mathbf{I})$ is the noise added to create $a^k$ from $a^0$.

The total training loss combines the diffusion loss with the modality regularization term:
\begin{equation}
\mathcal{L}_{total} = \mathcal{L}_{diff} + \mathcal{L}_{reg}
\end{equation}
where $\mathcal{L}_{reg}$ is the entropy regularization from AMAM that prevents uniform modality weighting.

For efficient inference, we adopt DDIM~\cite{song2020denoising} sampling to accelerate inference, requiring only 20 denoising steps to generate actions. We utilize a history of 3 observation frames to predict action chunks of horizon $H=8$ timesteps, then execute the first 6 actions before replanning. This chunked prediction reduces compounding errors while maintaining smooth trajectories necessary for stable multi-agent coordination~\cite{zhao2023learning}.

\subsection{Tactile-guided Grasping Strategy}

A critical challenge in multi-agent manipulation is achieving stable grasps despite visual uncertainties and occlusions. We observe that policies trained solely on standard grasping demonstrations often fail when deployed, with objects slipping during manipulation due to partial or misaligned contact. This occurs because visual perception alone cannot accurately determine optimal grasp configuration, especially when multiple agents create occlusions or when object surfaces have complex geometries.

To address this challenge, we propose a tactile-guided grasping strategy that leverages FSR feedback during data collection to teach the policy how to refine grasp contact dynamically. As illustrated in Fig.~\ref{fig:grasp_pattern}, our approach consists of two complementary data collection patterns:

\begin{figure*}
\centering
\captionsetup{font=footnotesize,labelsep=period}
\includegraphics[width=0.95\textwidth]{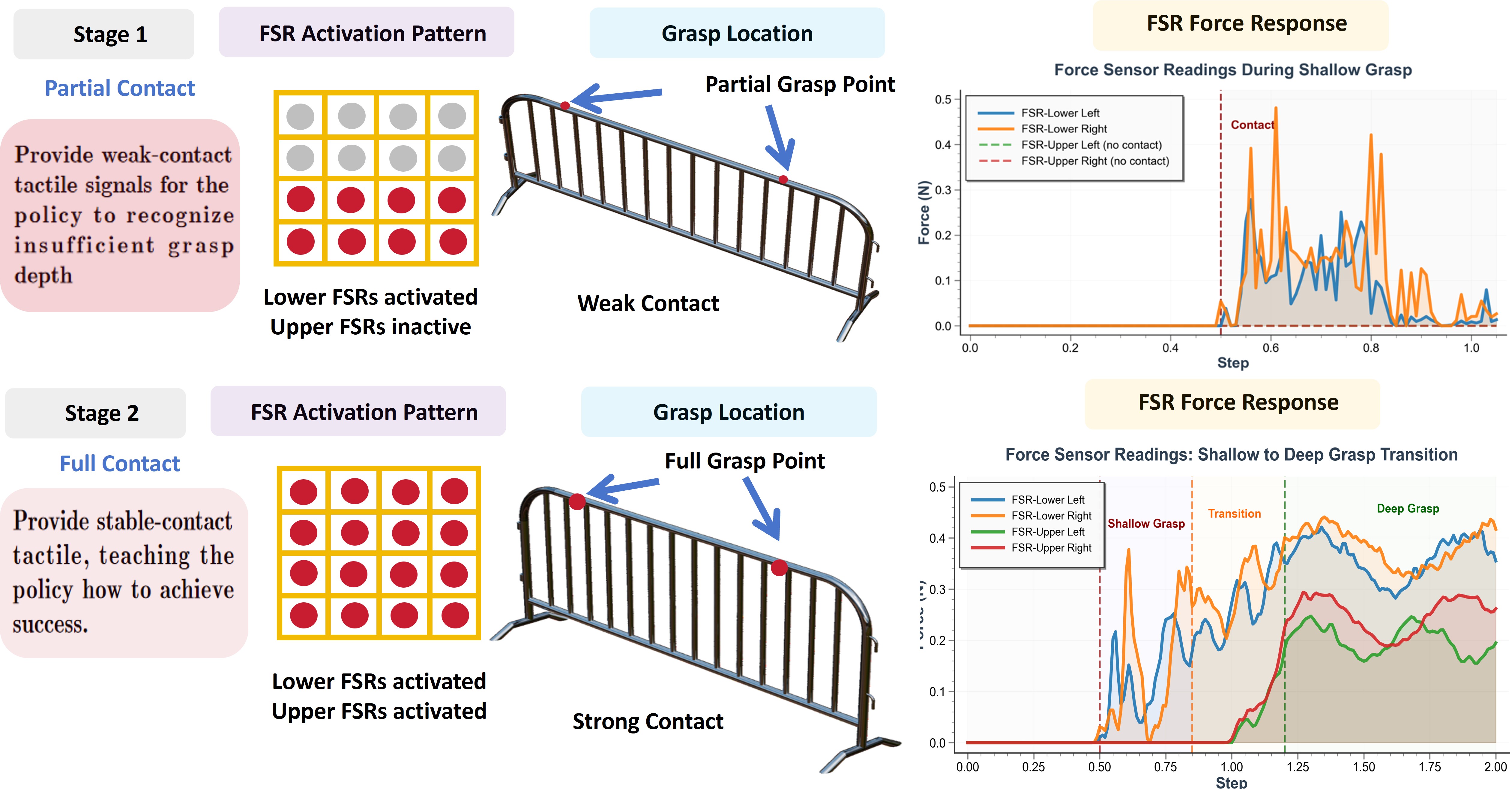}
\caption{Tactile-guided grasping strategy. (a) Standard full-contact grasp with complete tactile coverage across all FSR sensors. (b) Contact-refinement pattern where initial partial contact (limited FSR activation) triggers progressive grasping motion until a stable tactile signature is achieved. This teaches the policy to use tactile feedback for grasp refinement.}
\label{fig:grasp_pattern}
 \vspace{-0.1cm}
\end{figure*}

\textbf{(1) Standard full-contact grasps (70\% of data):} The gripper approaches the object and executes a stable grasp in which the FSR array exhibits broad activation and sufficient contact force across sensors. These demonstrations establish the nominal grasping behavior under accurate visual perception and serve as the baseline for successful manipulation.

\textbf{(2) Contact-refinement adjustments (30\% of data):} We deliberately begin with an incomplete grasp, indicated by weak or spatially partial FSR responses (e.g., activation concentrated on the lower sensors). The gripper then executes a corrective deepening motion along the approach direction while monitoring tactile feedback, continuing until broad sensor engagement and a more balanced force distribution are reached. This pattern teaches the policy to detect insufficient contact from tactile signatures and to perform refinement actions that recover a stable grasp.

%\textbf{Refinement variants.} We implement two refinement patterns: (i) \emph{release-regrasp}, where the gripper briefly releases, repositions deeper, and re-grasps, teaching explicit retry after failed contact; and (ii) \emph{in-grasp tightening}, where the gripper maintains contact and progressively closes from a partial closure to full contact, enabling continuous adjustment within a single grasp attempt.

Within the contact-refinement procedure, we employ two variants:  (1) \emph{release-regrasp}: an initial partial-contact grasp followed by gripper release, deeper repositioning, and re-grasping, which teaches the policy to recover from failed grasps via explicit retry; and  (2) \emph{in-grasp tightening}: a partially closed grasp with weak tactile readings followed by progressive closure to full contact, enabling continuous in-grasp adjustment without releasing the object.

%These data collection strategies enable the policy to learn a tactile-conditioned refinement behaviour. When deployed, if the model detects weak or partial tactile contact indicating insufficient coverage, it automatically triggers corrective grasping motions guided by the tactile encoder’s feedback. This closed-loop adjustment significantly improves grasp success rates, particularly in scenarios with visual ambiguity or when inter-agent occlusions compromise depth perception.

This protocol trains a tactile-conditioned refinement behavior. At test time, when the policy detects weak or spatially incomplete tactile contact, it triggers corrective grasp adjustments guided by the tactile encoder. The resulting closed-loop refinement improves grasp success, particularly under visual ambiguity or when inter-agent occlusions degrade depth perception.

\section{Experiments}

\subsection{Experimental Setup and Evaluation Tasks}

\begin{figure}
\centering
\captionsetup{font=footnotesize,labelsep=period}
\includegraphics[width=0.85\columnwidth]{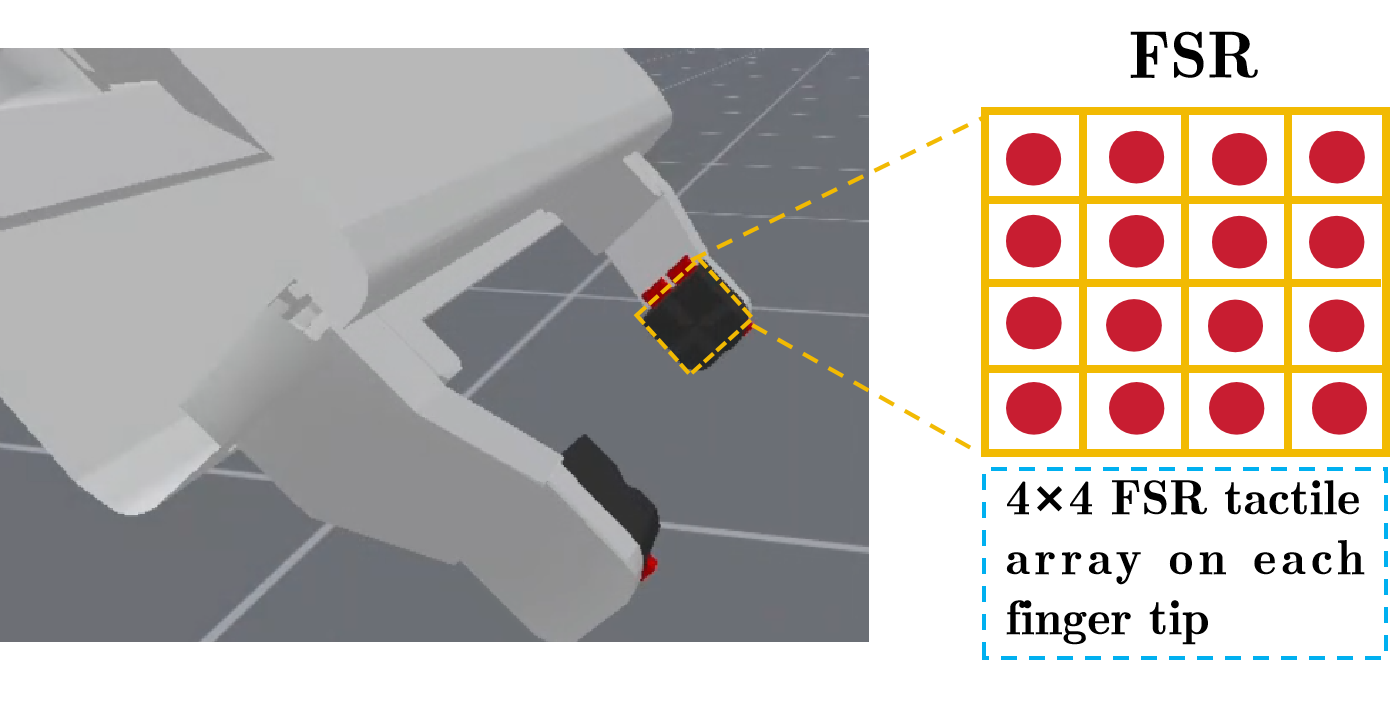}
\caption{FSR sensor integration on the Franka Panda gripper. Each 
FSR array is embedded within the compliant rubber fingertip, enabling spatially resolved tactile feedback during manipulation.}
\label{fig:fsr_setup}
 \vspace{-0.6cm}
\end{figure}

We evaluate our approach on seven multi-agent manipulation tasks using Franka Panda robots, each with 7 degrees of freedom plus gripper control. Each gripper is equipped with custom 4×4 FSR sensor arrays installed on the rubber tips of both fingers, as shown in Fig.~\ref{fig:fsr_setup}, providing 32 tactile readings per end-effector for fine-grained contact sensing.

\begin{figure}
\centering
\captionsetup{font=footnotesize,labelsep=period}
\includegraphics[width=\columnwidth]{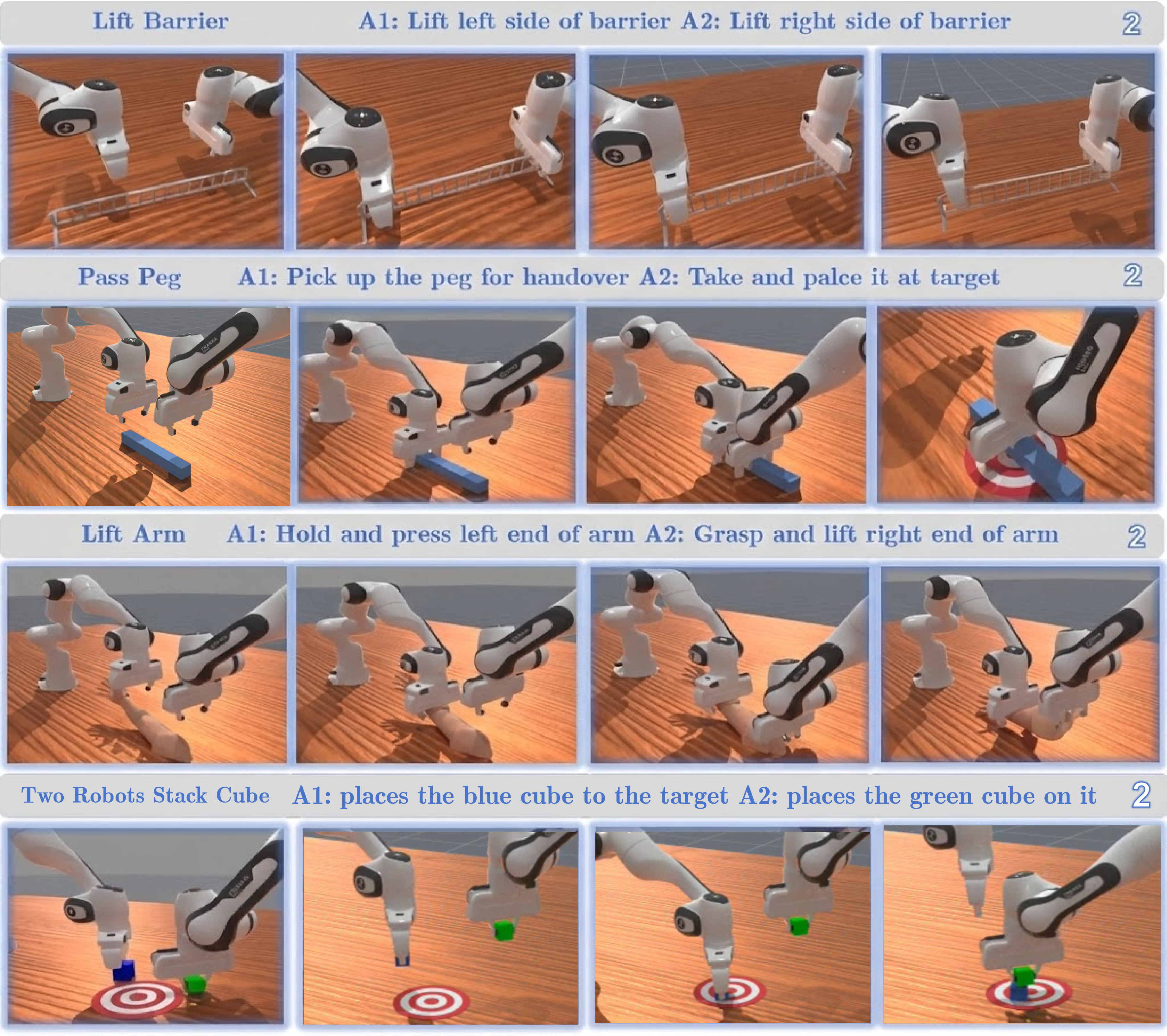}
\caption{Dual-arm manipulation tasks. From left to right: Lift Barrier, Pass Peg, Lift Arm, and Two Robots Stack Cube. Each task requires precise coordination between two agents with distinct roles specified through language instructions.}
\label{fig:tasks_two}
\end{figure}

\begin{figure}
\centering
\captionsetup{font=footnotesize,labelsep=period}
\includegraphics[width=\columnwidth]{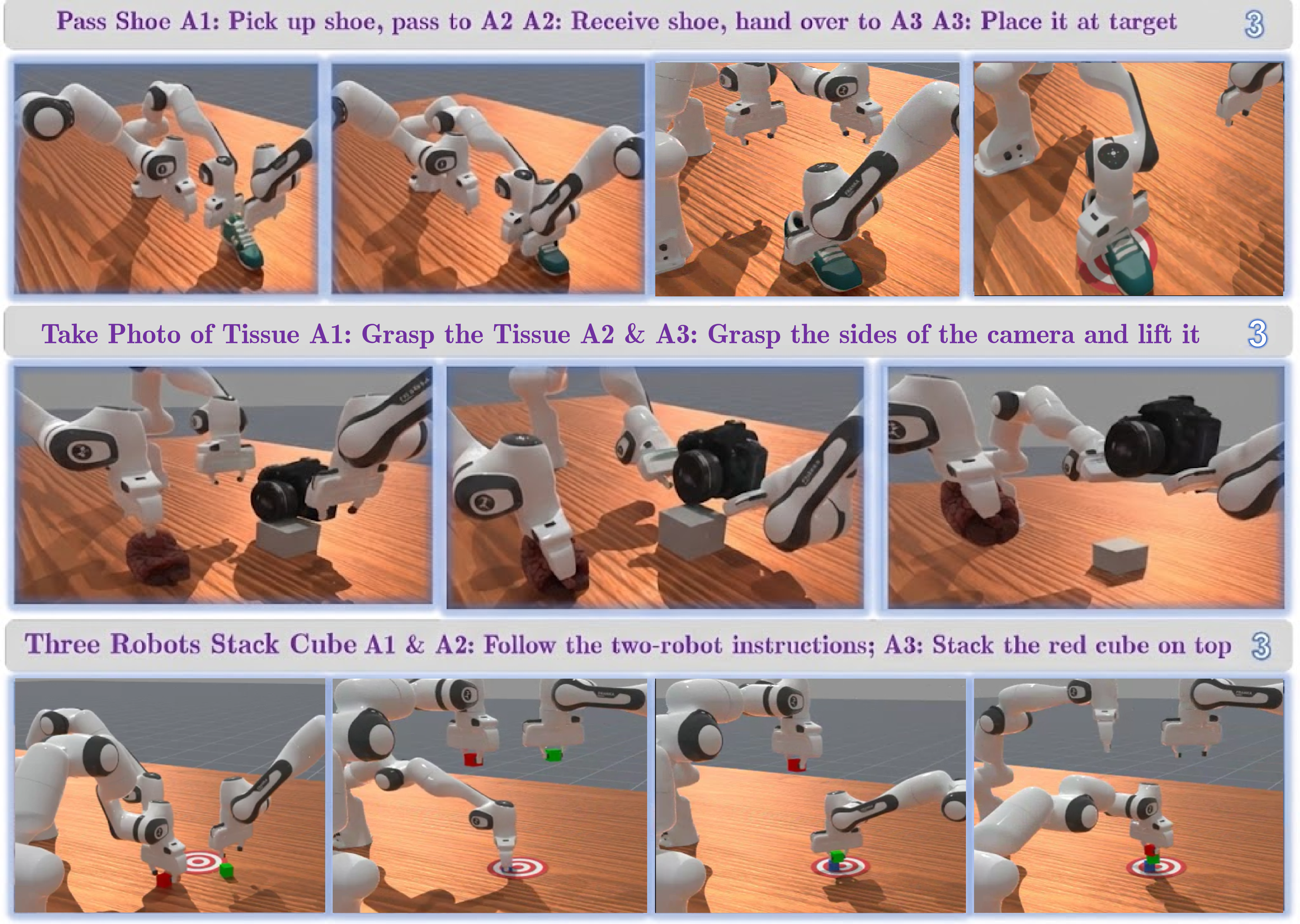}
\caption{Tri-arm manipulation tasks. From left to right: Pass Shoe, Take Photo of Tissue, and Three Robots Stack Cube. These tasks demonstrate scalability to three agents with complex spatial coordination requirements.}
\label{fig:tasks_three}
 \vspace{-0.5cm}
\end{figure}

Our benchmark consists of four dual-arm and three tri-arm manipulation tasks adapted from ManiSkill~\cite{tao2024maniskill3} and RoboFactory~\cite{qin2025robofactory}, as illustrated in Fig.~\ref{fig:tasks_two} and Fig.~\ref{fig:tasks_three}. The dual-arm tasks include \textit{Lift Barrier}, \textit{Pass Peg}, \textit{Lift Arm}, and \textit{Two Robots Stack Cube}, while the tri-arm tasks comprise \textit{Pass Shoe}, \textit{Take Photo of Tissue}, and \textit{Three Robots Stack Cube}. Notably, the \textit{Pass Peg} and \textit{Pass Shoe} tasks are specifically designed with close-proximity handover sequences where agents operate within overlapping workspaces, creating challenging scenarios that test our graph-based collision avoidance mechanism. Besides, each agent receives task-specific language instructions.

For each task, we collect expert demonstrations using motion planners with randomized object positions and orientations. We evaluate all methods using two data regimes: 50 and 150 demonstrations per agent, and measure the performance by success rate.

\subsection{Baseline Methods and Comparative Results}

We evaluate our approach against state-of-the-art imitation learning methods for robotic manipulation. As baselines, we selected two diffusion-based policies: (1) Diffusion Policy (DP)~\cite{chi2023diffusion}, which utilizes RGB images as observations to generate actions via diffusion models; and (2) 3D Diffusion Policy (DP3)~\cite{ze20243d}, which replaces image inputs in Diffusion Policy with point clouds for improved 3D spatial reasoning. Additionally, we compare with (3) Flow Policy~\cite{zhang2025flowpolicy}, which leverages flow matching~\cite{lipman2022flow} with point cloud observations, offering faster inference through straight trajectory generation while maintaining generation quality comparable to diffusion models.

\begin{figure}
\centering
\captionsetup{font=footnotesize,labelsep=period}
\includegraphics[width=0.82\columnwidth]{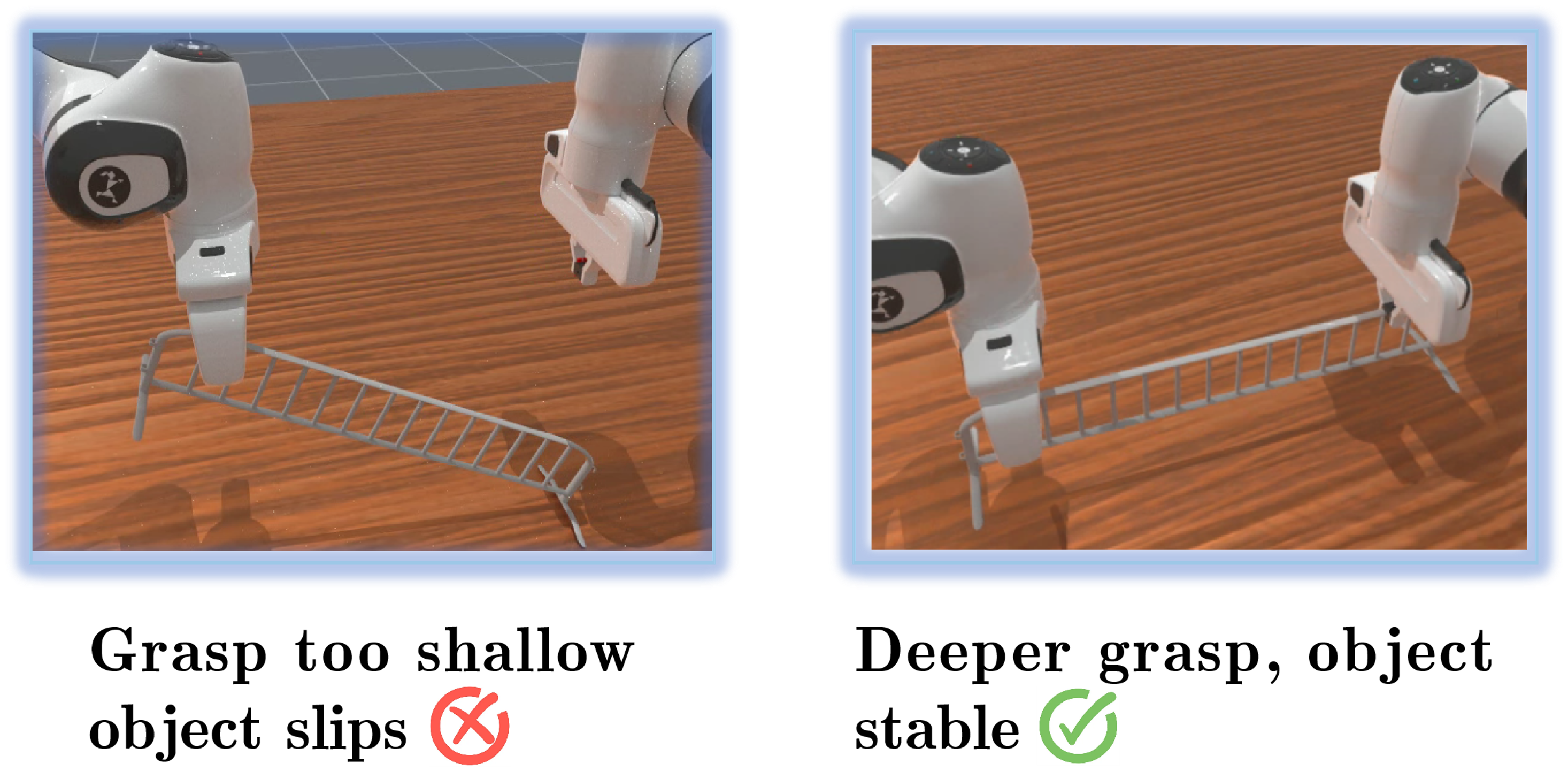}
\caption{Left: Shallow grasp causes slipping. Right: Tactile feedback enables more refine, stable grasping.}
\label{fig:grasp_comparison}
 \vspace{-0.3cm}
\end{figure}

\begin{figure}
\centering
\captionsetup{font=footnotesize,labelsep=period}
\includegraphics[width=0.85\columnwidth]{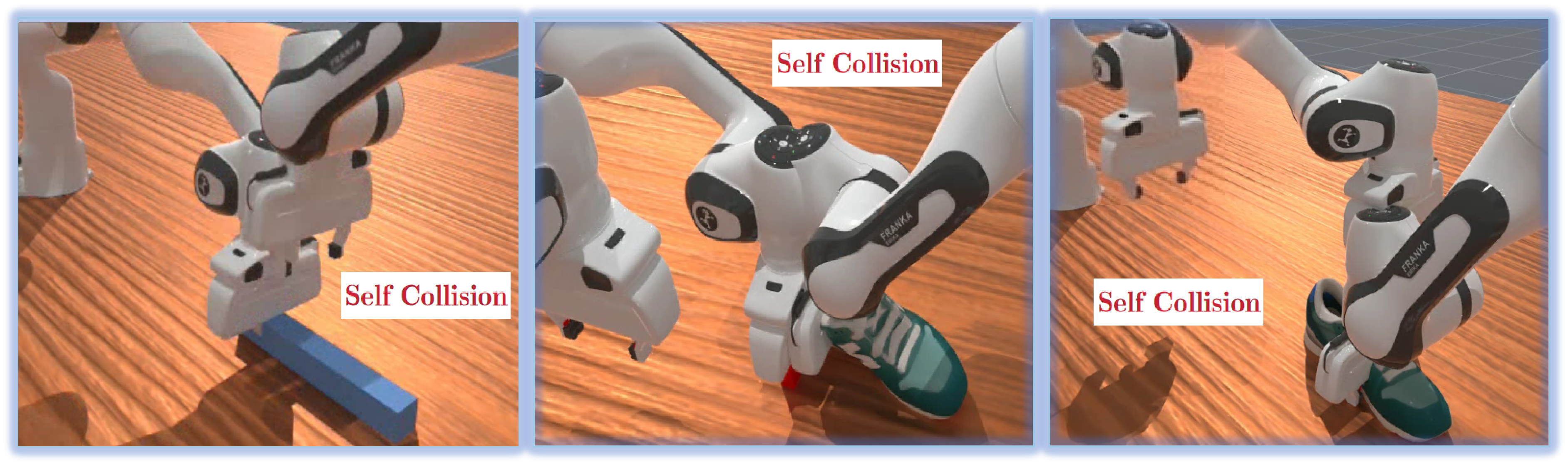}
\caption{Self-collisions during multi-agent manipulation without graph-based coordination.}
\label{fig:collision}
 \vspace{-0.5cm}
\end{figure}

\begin{table*}[t]
\centering
\captionsetup{font=footnotesize,labelsep=period}
\caption{Success rates (\%) on multi-agent manipulation tasks with different numbers of demonstrations}
\label{tab:main_results}
\begin{tabular}{llcccc|cccc}
\toprule
& & \multicolumn{4}{c|}{50 Demonstrations} & \multicolumn{4}{c}{150 Demonstrations} \\
\cmidrule(lr){3-6} \cmidrule(lr){7-10}
Agents & Task & DP & DP3 & Flow Policy & Ours & DP & DP3 & Flow Policy & Ours \\
\midrule
\multirow{4}{*}{Two-Agent} 
& Lift Barrier & 25 & 33 & 36 & \textbf{55} & 68 & 77 & 75 & \textbf{92} \\
& Pass Peg & 28 & 41 & 35 & \textbf{52} & 42 & 50 & 53 & \textbf{74} \\
& Lift Arm & 17 & 26 & 24 & \textbf{41} & 36 & 62 & 58 & \textbf{78} \\
& Two Robots Stack Cube & 14 & 19 & 22 & \textbf{24} & 21 & 37 & 40 & \textbf{44} \\
\midrule
\multirow{3}{*}{Three-Agent}
& Pass Shoe & 9 & 15 & 13 & \textbf{36} & 16 & 30 & 28 & \textbf{37} \\
& Take Photo of Tissue & 7 & 17 & 19 & \textbf{31} & 19 & 33 & 31 & \textbf{42} \\
& Three Robots Stack Cube & 8 & 19 & 21 & \textbf{25} & 21 & 28 & 32 & \textbf{35} \\
\midrule
\multicolumn{2}{l}{Average (Two-Agent)} & 21.0 & 29.8 & 29.3 & \textbf{43.0} & 41.8 & 56.5 & 56.5 & \textbf{72.0} \\
\multicolumn{2}{l}{Average (Three-Agent)} & 8.0 & 17.0 & 17.7 & \textbf{30.7} & 18.7 & 30.3 & 30.3 & \textbf{38.0} \\
\multicolumn{2}{l}{Overall Average} & 15.4 & 24.3 & 24.3 & \textbf{37.6} & 31.7 & 45.1 & 45.4 & \textbf{57.1} \\
\bottomrule
\end{tabular}
\end{table*}

Table~\ref{tab:main_results} presents the comparative results across all tasks. Our method consistently outperforms all baselines in both data regimes, with particularly significant improvements in the low-data setting (50 demonstrations). In the 50-demonstration regime, ADM-DP achieves an average improvement of 13.3\% over the best baseline, demonstrating superior data efficiency. This advantage is maintained with 150 demonstrations, where we achieve 57.1\% overall success rate compared to 45.4\% for the next best method.

The performance gains are particularly pronounced in tasks requiring substantial tactile feedback. For \textit{Lift Barrier} and \textit{Lift Arm}, which involve precise force control for stable lifting, our tactile-guided approach achieves 92\% and 78\% success rates respectively with 150 demonstrations, significantly outperforming DP3's 77\% and 62\%. The tactile modality enables our policy to detect and correct unstable grasps that vision-only methods fail to identify.
For handover tasks (\textit{Pass Peg} and \textit{Pass Shoe}), which demand close-proximity coordination, our graph-based collision encoding provides critical spatial awareness. Baseline methods frequently collide when agents operate in overlapping workspaces; in contrast, ADM-DP maintains safe separation by leveraging TCP-based graph features. As a result, ADM-DP achieves 74\% success on \textit{Pass Peg}, compared to 53\% for Flow Policy.

Notably, even on vision-dominant tasks such as \textit{Stack Cube}, our enhanced visual encoder—fusing RGB and point-cloud features via FiLM yields more reliable 3D scene understanding and consistently outperforms single-modality baselines. These gains across heterogeneous task requirements support the effectiveness of adaptive fusion: multimodal tasks (e.g., \textit{Pass Peg}, which requires vision, tactile feedback, and inter-agent spatial reasoning) benefit most when AMAM emphasizes complementary signals, whereas simpler tasks improve when AMAM down-weights irrelevant modalities to reduce noise.
Finally, the results highlight the increased difficulty of scaling to three-agent settings, where all methods exhibit performance degradation. Nevertheless, ADM-DP preserves the largest relative advantage, achieving nearly twice the success rate of DP in the three-agent regime, underscoring the scalability benefits of our design.

\subsection{Ablation Studies}

To analyze the contribution of each component in our architecture, we conduct ablation studies by systematically removing key modules. Table~\ref{tab:ablation} presents the results with 150 demonstrations, where ADM-NoPC removes point cloud input (using only RGB), ADM-NoTact removes tactile sensing, ADM-NoGraph removes the graph-based collision encoding, and ADM-NoAM removes the AMAM module (using simple concatenation instead).

% \begin{figure}
% \centering
% \captionsetup{font=footnotesize,labelsep=period}
% \includegraphics[width=0.85\columnwidth]{fig/grasp_fail1.png}
% \caption{Left: Shallow grasp causes slipping. Right: Tactile feedback enables more refine, stable grasping.}
% \label{fig:grasp_comparison}
%  \vspace{-0.3cm}
% \end{figure}

\begin{table*}[t]
\centering
\captionsetup{font=footnotesize,labelsep=period}
\caption{Ablation study results showing success rates (\%) with 150 demonstrations}
\label{tab:ablation}
\begin{tabular}{llccccc}
\toprule
Agents & Task & ADM-DP & ADM-NoPC & ADM-NoTact & ADM-NoGraph & ADM-NoAM \\
\midrule
\multirow{4}{*}{Two-Agent} 
& Lift Barrier & \textbf{92} & 83 & 78 & 89 & 84 \\
& Pass Peg & \textbf{74} & 64 & 70 & 67 & 68 \\
& Lift Arm & \textbf{78} & 61 & 68 & 76 & 73 \\
& Two Robots Stack Cube & \textbf{44} & 26 & 41 & 42 & 41 \\
\midrule
\multirow{3}{*}{Three-Agent}
& Pass Shoe & \textbf{37} & 32 & 30 & 28 & 28 \\
& Take Photo of Tissue & \textbf{42} & 31 & 36 & 40 & 37 \\
& Three Robots Stack Cube & \textbf{35} & 24 & 33 & 34 & 33 \\
\midrule
\multicolumn{2}{l}{Average} & \textbf{57.4} & 45.9 & 50.9 & 53.7 & 52.0 \\
\bottomrule
\end{tabular}
 \vspace{-0.5cm}
\end{table*}

% \begin{figure}
% \centering
% \captionsetup{font=footnotesize,labelsep=period}
% \includegraphics[width=0.85\columnwidth]{fig/collision1.png}
% \caption{Self-collisions during multi-agent manipulation without graph-based coordination.}
% \label{fig:collision}
%  \vspace{-0.3cm}
% \end{figure}

The ablation results reveal distinct patterns in component importance across different tasks. Removing point cloud input (ADM-NoPC) causes the most significant performance degradation overall (11.5\% drop), particularly affecting visually-demanding tasks like \textit{Two Robots Stack Cube} where success rate drops from 44\% to 26\%. This validates our enhanced visual encoding strategy that leverages both RGB semantics and point cloud geometry.

Tactile feedback is critical for tasks that demand grasp stability and force regulation. On \textit{Lift Barrier}, removing tactile input leads to a 14\% absolute drop in success rate (92\% $\rightarrow$ 78\%), as the policy can no longer reliably detect insufficient contact or execute corrective grasp refinements. Likewise, performance on \textit{Lift Arm} decreases by 10\%, further supporting the effectiveness of our tactile-guided grasping strategy for mitigating unstable grasps. As illustrated in Fig.~\ref{fig:grasp_comparison}, shallow grasps can induce slipping when tactile feedback is absent, due to the lack of signal to identify inadequate contact and adjust grasp depth accordingly.
%Tactile sensing proves crucial for tasks requiring stable grasping and force control. \textit{Lift Barrier} shows a 14\% performance drop without tactile feedback (92\% to 78\%), as the policy loses the ability to detect and correct unstable grasps. Similarly, \textit{Lift Arm} degrades by 10\%, confirming that our tactile-guided grasping strategy effectively addresses grasp stability challenges. Fig.~\ref{fig:grasp_comparison} shows how shallow grasps cause object slipping without tactile feedback to detect insufficient contact.

The graph module's importance is most evident in close-proximity coordination tasks. \textit{Pass Peg} and \textit{Pass Shoe} show 7\% and 9\% drops respectively without graph encoding, as agents lose spatial awareness and experience more frequent collisions during handovers. The graph features enable implicit coordination through shared TCP positions, critical for safe multi-agent interaction. Fig.~\ref{fig:collision} shows typical self-collisions that occur without graph-based spatial awareness.

\begin{figure}
\centering
\captionsetup{font=footnotesize,labelsep=period}
\includegraphics[width=0.95\linewidth
]{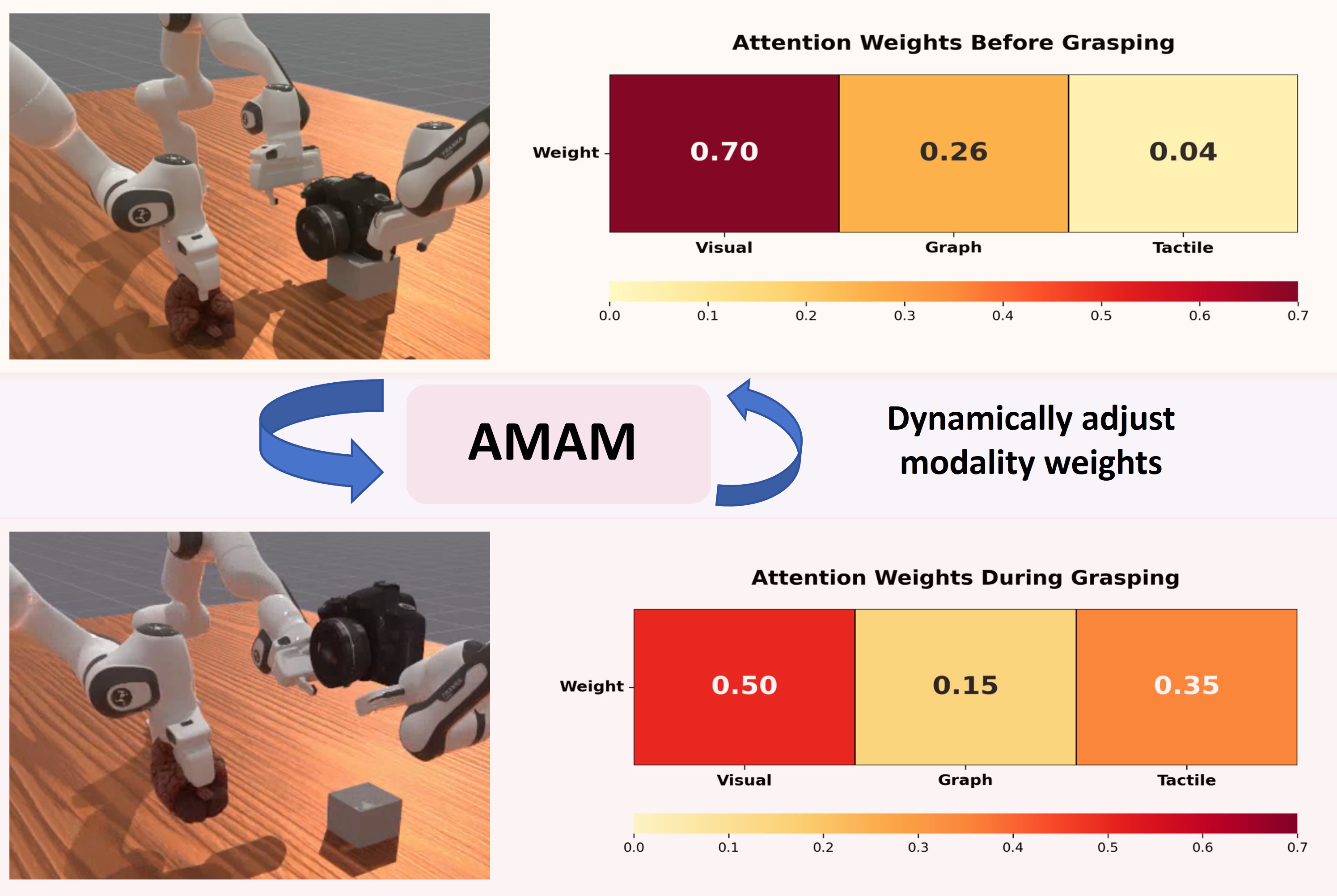}
\caption{Dynamic modality weight adaptation by AMAM.}
\label{fig:weight_change}
 \vspace{-0.5cm}
\end{figure}

Removing AMAM (ADM-NoAM) results in a consistent performance decrease across all tasks (5.4\% average drop), with the impact correlating strongly with task complexity. Multi-modal tasks show the largest degradation: \textit{Pass Shoe}, which requires coordinated use of vision, tactile, and graph features, drops 9\% (from 37\% to 28\%), while simpler vision-dominant tasks like \textit{Stack Cube} only drop 3\%. This pattern confirms that AMAM's dynamic weighting becomes increasingly valuable as tasks demand integration of multiple sensory streams. Rather than using fixed fusion weights that may over-emphasize irrelevant modalities, AMAM adaptively allocates attention based on task context, preventing noise from unused sensors while ensuring critical modalities receive appropriate emphasis when needed. %Figure~\ref{fig:weight_change} visualizes this dynamic adaptation, showing AMAM automatically shifting from vision-dominant weights before grasping to balanced multi-modal attention during grasp.
Fig.~\ref{fig:weight_change} visualizes this dynamic adaptation, showing AMAM automatically shifting from vision-dominant weights (0.70) before grasping to balanced multi-modal attention during grasp, where tactile increases to 0.35 for contact feedback while vision decreases to 0.50, demonstrating context-aware modality prioritization.

These ablations show that each component targets a distinct bottleneck in multi-agent manipulation: enhanced vision improves perception, tactile feedback stabilizes grasps, graph encoding reduces collisions, and AMAM enables context-dependent fusion.

%The complementary nature of these modules enables our method's robust performance across diverse coordination requirements.

% \begin{figure}
% \centering
% \captionsetup{font=footnotesize,labelsep=period}
% \includegraphics[width=0.98\columnwidth]{fig/weight.png}
% \caption{Dynamic modality weight adaptation by AMAM during task execution.}
% \label{fig:weight_change}
%  \vspace{-0.3cm}
% \end{figure}

\section{Conclusions and Future Work}

We presented ADM-DP, a Multimoal Diffusion Policy for multi-agent robotic manipulation that addresses key challenges in coordination, grasping stability, and collision avoidance. Our approach combines four key innovations: (1) enhanced visual encoding through FiLM-based fusion of RGB and point cloud features, (2) tactile-guided grasping strategy with spatial encoding that enables dynamic grasp adjustment, (3) graph-based collision awareness through shared TCP positions, and (4) AMAM that dynamically allocates attention across modalities based on task context. Through extensive experiments on seven multi-agent tasks involving two and three robots, we demonstrated that ADM-DP significantly outperforms state-of-the-art baselines, achieving 57.1\% average success rate compared to 45.4\% for the next best method. Our ablation studies confirmed that each component addresses specific challenges, with performance gains most pronounced in tasks requiring multiple sensory modalities. The decoupled training paradigm enables efficient scaling to multiple agents while maintaining modularity for system deployment. 

Future work will focus on real robot deployment to validate ADM-DP's effectiveness beyond simulation, addressing sensor noise and real-time constraints. We also plan to explore advanced tactile sensing technologies like vision-based tactile sensors for more complex manipulation tasks requiring delicate force control.

\bibliographystyle{IEEEtran}
\bibliography{reference}

\end{document}